\documentclass[runningheads]{llncs}
\usepackage{graphicx}

\usepackage{tikz}
\usepackage{comment}
\usepackage{amsmath,amssymb} 
\usepackage{color}

\usepackage{cite}
\usepackage{subcaption}
\usepackage{graphicx}
\usepackage{multirow}

\definecolor{_yellow}{RGB}{255,217,50}
\definecolor{_darkblue}{RGB}{1,25,147}
\definecolor{_red}{RGB}{238,34,12}
\definecolor{_pink}{RGB}{255,64,255}
\definecolor{_orange}{RGB}{255,147,0}
\definecolor{_blue}{RGB}{4,51,255}
\definecolor{_green}{RGB}{97,216,54}
\definecolor{_purple}{RGB}{122,129,255}
\definecolor{_skyblue}{RGB}{86,193,255}
\definecolor{_coral}{RGB}{255,100,78}
\definecolor{_violet}{RGB}{148,55,250}

\newcommand{\rankfirst}[1]{\textbf{\textcolor{red}{#1}}}
\newcommand{\ranksecond}[1]{\textbf{\textcolor{blue}{#1}}}

\newcommand{\best}[1]{\textbf{#1}}

\usepackage[pagebackref,breaklinks,colorlinks]{hyperref}

\usepackage[accsupp]{axessibility}  

\newcommand{\beginappendix}{%
        \setcounter{section}{0}
        \renewcommand\thesection{\Alph{section}}
        \setcounter{table}{0}
        \renewcommand{\thetable}{A\arabic{table}}%
        \setcounter{figure}{0}
        \renewcommand{\thefigure}{A\arabic{figure}}%
     }

\begin{document}
\pagestyle{headings}
\mainmatter
\def\ECCVSubNumber{xxxx}  

\title{Joint Feature Learning and Relation Modeling for Tracking: A One-Stream Framework} 

\titlerunning{Joint Feature Learning and Relation Modeling for Tracking}
%
\author{Botao Ye\inst{1,2}\and
Hong Chang\inst{1,2} \and
Bingpeng Ma\inst{2} \and \\
Shiguang Shan\inst{1,2} \and
Xilin Chen\inst{1,2}
}

%
\authorrunning{B. Ye, H. Chang et al.}
%

\institute{Key Lab of Intelligent Information Processing of Chinese Academy of Sciences (CAS), Institute of Computing Technology, CAS, Beijing, 100190, China \and
University of Chinese Academy of Sciences, Beijing, 100049, China \\
\email{botao.ye@vipl.ict.ac.cn, changhong@ict.ac.cn, bpma@ucas.ac.cn,  \\ \{sgshan, xlchen\}@ict.ac.cn}
}

\maketitle

\begin{abstract}
The current popular two-stream, two-stage tracking framework extracts the template and the search region features separately and then performs relation modeling, thus the extracted features lack the awareness of the target and have limited target-background discriminability.
To tackle the above issue, we propose a novel \emph{one-stream tracking} (OSTrack) framework that unifies feature learning and relation modeling by bridging the template-search image pairs with bidirectional information flows. In this way, discriminative target-oriented features can be dynamically extracted by mutual guidance.
Since no extra heavy relation modeling module is needed and the implementation is highly parallelized, the proposed tracker runs at a fast speed. 
To further improve the inference efficiency, an in-network candidate early elimination module is proposed based on the strong similarity prior calculated in the one-stream framework.
As a unified framework, OSTrack achieves state-of-the-art performance on multiple benchmarks, in particular, it shows impressive results on the one-shot tracking benchmark GOT-10k, \ie, achieving 73.7\% AO, improving the existing best result (SwinTrack) by 4.3\%. Besides, our method maintains a good performance-speed trade-off and shows faster convergence. The code and models are available at \url{https://github.com/botaoye/OSTrack}.


\end{abstract}

\setlength{\floatsep}{1pt}
\setlength{\textfloatsep}{16pt}

\section{Introduction}
Visual object tracking (VOT) aims at localizing an arbitrary target in each video frame, given only its initial appearance. The \emph{continuously changing} and \emph{arbitrary} nature of the target poses a challenge to learn a target appearance model that can effectively discriminate the specified target from the background. Current mainstream trackers typically address this problem with a common \emph{two-stream} and \emph{two-stage} pipeline, which means that the features of the template and the search region are separately extracted (two-stream), and the whole process is divided into two sequential steps: feature extraction and relation modeling (two-stage). Such a natural pipeline employs the strategy of ``divide-and-conquer'' and achieves remarkable success in terms of tracking performance.

However, the separation of feature extraction and relation modeling suffers from the following limitations.
Firstly, the feature extracted by the vanilla two-stream two-stage framework is unaware of the target. In other words, the extracted feature for each image is determined after off-line training, since there is no interaction between the template and the search region. This is against with the continuously changing and arbitrary nature of the target, leading to limited target-background discriminative power.
On some occasions when the category of the target object is not involved in the training dataset (\ie, one-shot tracking),  
the above problems are particularly serious.
Secondly, the two-stream, two-stage framework is vulnerable to the performance-speed dilemma.
According to the computation burden of the feature fusion module, two different strategies are commonly utilized. The first type, shown in Fig~\ref{fig:taxonomy}(a), simply adopts one single operator like cross-correlation~\cite{siamfc, siamrpn} or discriminative correlation filter~\cite{atom, dimp}, which is efficient but less effective since the simple linear operation leads to discriminative information loss~\cite{transt}. The second type, shown in Fig~\ref{fig:taxonomy}(b), addresses the information loss by complicated non-linear interaction (Transformer~\cite{transformer}), but is less efficient due to a large number of parameters and the use of iterative refinement (\eg, for each search image, STARK-S50~\cite{stark} takes 7.5 ms for the feature extraction and 14.1 ms for relation modeling on an RTX2080Ti GPU).

\begin{figure}[t]
	\centering
	\begin{minipage}[c]{.42\textwidth}
		\begin{flushleft}
					\caption{A comparison of AO and speed of state-of-the-art trackers on GOT-10k under one-shot setting.  Our OSTrack-384 sets a new SOTA of 73.7\% AO on GOT-10k, showing impressive one-shot tracking performance. OSTrack-256 runs at 105.4 FPS while still outperforming all previous trackers.}\label{fig:efficiency}

		\end{flushleft}
	\end{minipage}%
	\hspace{0.1em}
	\begin{minipage}[c]{.53\textwidth}
		\centering
		\begin{flushright}
			\includegraphics[width=1\textwidth,height=0.74\textwidth]{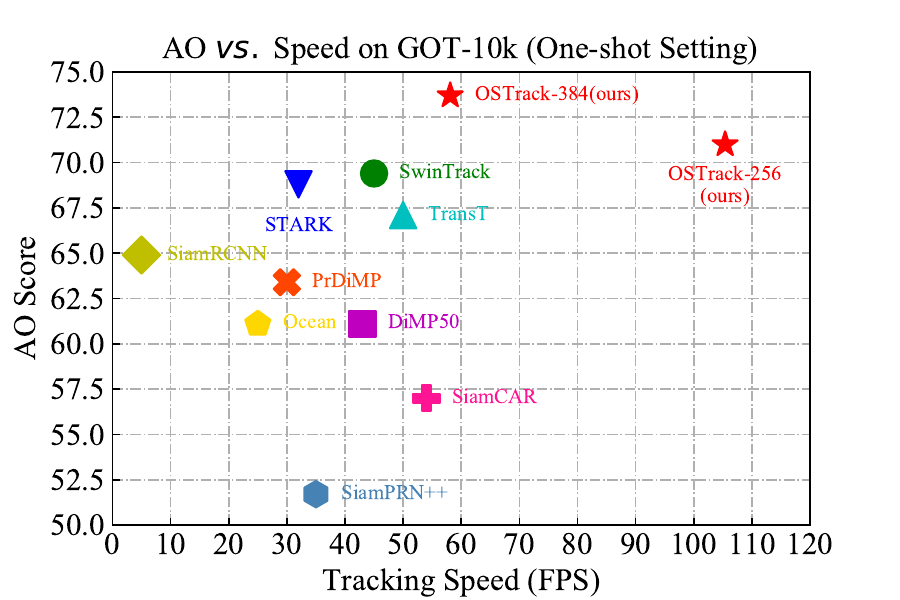}
		\end{flushright}
	\end{minipage}
\end{figure}

In this work, we set out to address the aforementioned problems via a unified \emph{one-stream one-stage} tracking framework. The core insight of the one-stream framework is to bridge a free information flow between the template and search region at the early stage (\ie, the raw image pair), thus extracting target-oriented features and avoiding the loss of discriminative information. Specifically, we concatenate the flattened template and search region and feed them into staked self-attention layers~\cite{transformer} (widely used Vision Transformer (ViT)~\cite{vit} is chosen in our implementation), and the produced search region features can be directly used for target classification and regression without further matching. The staked self-attention operations enable iteratively feature matching between the template and the search region, thus allowing mutual guidance for target-oriented feature extraction. Therefore, both template and search region features can be extracted dynamically with strong discriminative power.
Additionally, the proposed framework achieves a good balance between performance and speed because the concatenation of the template and the search region makes the one-stream framework highly parallelizable and does not require additional heavy relational modeling networks.

Moreover, the proposed one-stream framework provides a strong prior about the similarity of the target and each part of the search region (\ie candidates) as shown in Fig.~\ref{fig:vis_attn}, which means that the model can identify background regions even at the early stage. This phenomenon verifies the effectiveness of the one-stream framework and motivates us to propose an in-network \emph{early candidate elimination} module for progressively identifying and discarding the candidates belonging to the background in a timely manner. The proposed candidate elimination module not only significantly boosts the inference speed, but also avoids the negative impact of uninformative background regions on feature matching.

Despite its simple structure, the proposed trackers achieve impressive performance and set a new state-of-the-art (SOTA) on multiple benchmarks. Moreover, it maintains adorable inference efficiency and shows faster convergence compared to SOTA Transformer based trackers. As shown in Fig.~\ref{fig:efficiency}, our method achieves a good balance between the accuracy and inference speed.

The main contributions of this work are three-fold: 
(1) We propose a simple, neat, and effective one-stream, one-stage tracking framework by combining the feature extraction and relation modeling.
(2) Motivated by the prior of the early acquired similarity score between the target and each part of the search region, an in-network early candidate elimination module is proposed for decreasing the inference time. 
(3) We perform comprehensive experiments to verify that the one-stream framework outperforms the previous SOTA two-stream trackers in terms of performance, inference speed, and convergence speed. The resulting tracker OSTrack sets a new state-of-the-art on multiple tracking benchmarks.
 
\section{Related Work}
In this section, we briefly review different tracking pipelines,  as well as the adaptive inference methods related to our early candidate elimination module.

\begin{figure}[t]
\centering
\includegraphics[width=0.88\columnwidth, keepaspectratio]{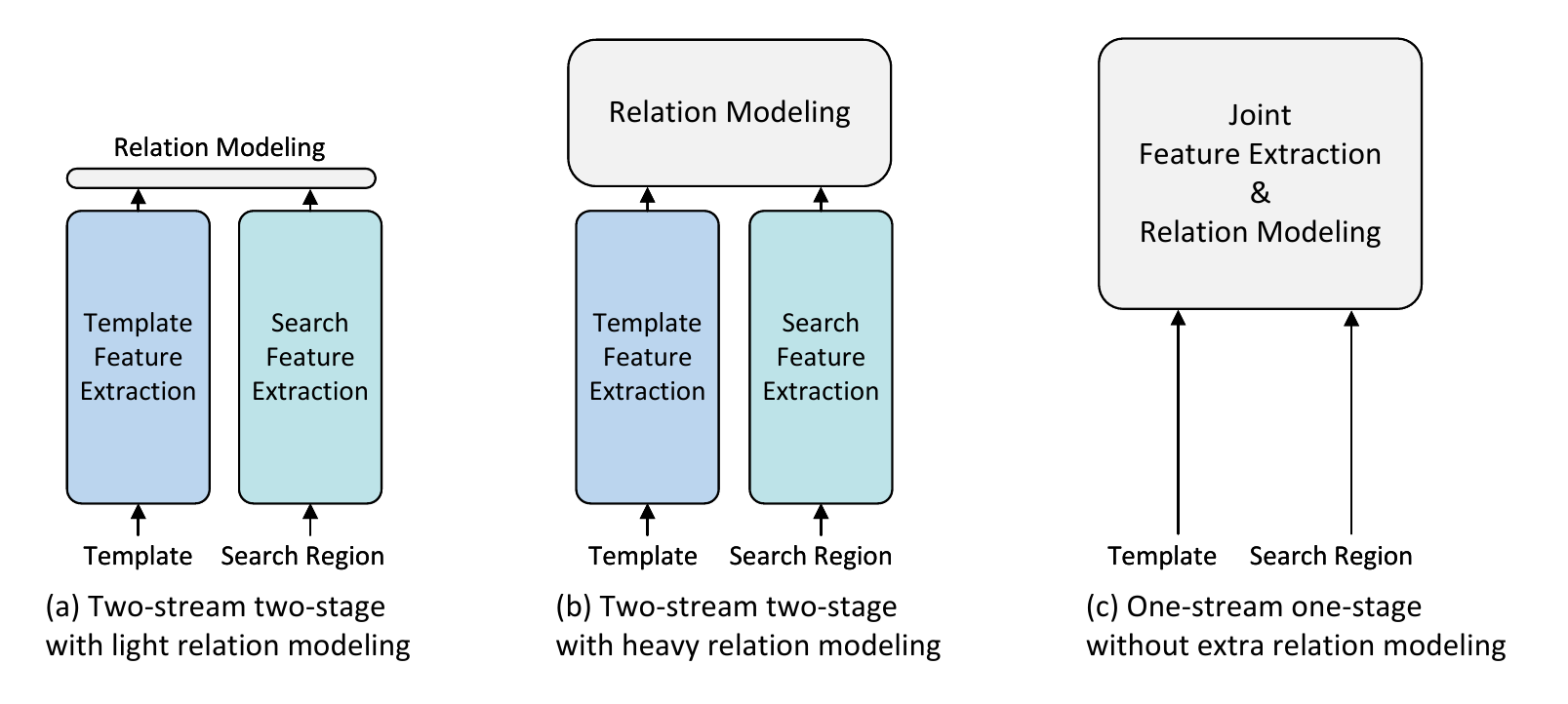}
\caption{Three different taxonomies of tracking pipeline. The height of each rectangular represents the relative model size.}
\label{fig:taxonomy}
\end{figure}

\textbf{Tracking Pipelines.}
Based on the different computational burdens of feature extraction and relation modeling networks, we compare our method with two different two-stream two-stage archetypes in Fig.~\ref{fig:taxonomy}. Earlier Siamese trackers~\cite{siamfc, siamrpn, ocean} and discriminative trackers~\cite{atom, dimp} belong to Fig.~\ref{fig:taxonomy}(a). They first extract the features of the template and the search region separately by a CNN backbone~\cite{alexnet, resnet}, which shares the same structure and parameters. Then, a lightweight relation modeling network (\eg, the cross-correlation layer~\cite{siamfc, siamrpn++} in Siamese trackers and correlation filter~\cite{mosse, kcf} in discriminative trackers) takes responsibility to fuse these features for the subsequent state estimation task. However, the template feature cannot be adjusted according to the search region feature in these methods. Such a shallow and unidirectional relation modeling strategy may be insufficient for information interaction.
ARecently, stacked Transformer layers~\cite{transformer} are introduced for better relation modeling. These methods belong to Fig.~\ref{fig:taxonomy}(b) where the relation modeling module is relatively heavy and enables bi-directional information interaction. TransT~\cite{transt} proposes to stack a series of self-attention and cross-attention layers for iterative feature fusion. STARK~\cite{stark} concatenates the pre-extracted template and search region features and feeds them into multiple self-attention layers. The bi-directional heavy structure brings performance gain but inevitably slows down the inference speed.
Differently, our one-stream one-stage design belongs to Fig.~\ref{fig:taxonomy}(c). For the first time, we seamlessly combine feature extraction and relation modeling into a unified pipeline. The proposed method provides free information flow between the template and search region with minor computation costs. It not only generates target-oriented features by mutual guidance but also is efficient in terms of both training and testing time.

\textbf{Adaptive Inference.}
Our early candidate elimination module can be seen as a progressive process of adaptively discarding potential background regions based on the similarity between the target and the search region. One related topic is the adaptive inference~\cite{dynamicvit, evit, adavit} in vision transformers, which is proposed to accelerate the computation of ViT. DynamicViT~\cite{dynamicvit} trains extra control gates with the Gumbel-softmax trick to discard tokens during inference. Instead of directly discarding non-informative tokens, EViT~\cite{evit} fuses them to avoid potential information loss. These works are tightly coupled with the classification task and are therefore not suitable for tracking. Instead, we treat each token as a target candidate and then discard the candidates that are least similar to the target by means of a free similarity score calculated by the self-attention operation. To the best of our knowledge, this is the first work that attempts to eliminate potential background candidates within the tracking network.

\section{Method}
\label{sec:method}
\begin{figure}[t]
\centering
\setlength{\abovecaptionskip}{0.05cm}
\includegraphics[width=\columnwidth, keepaspectratio]{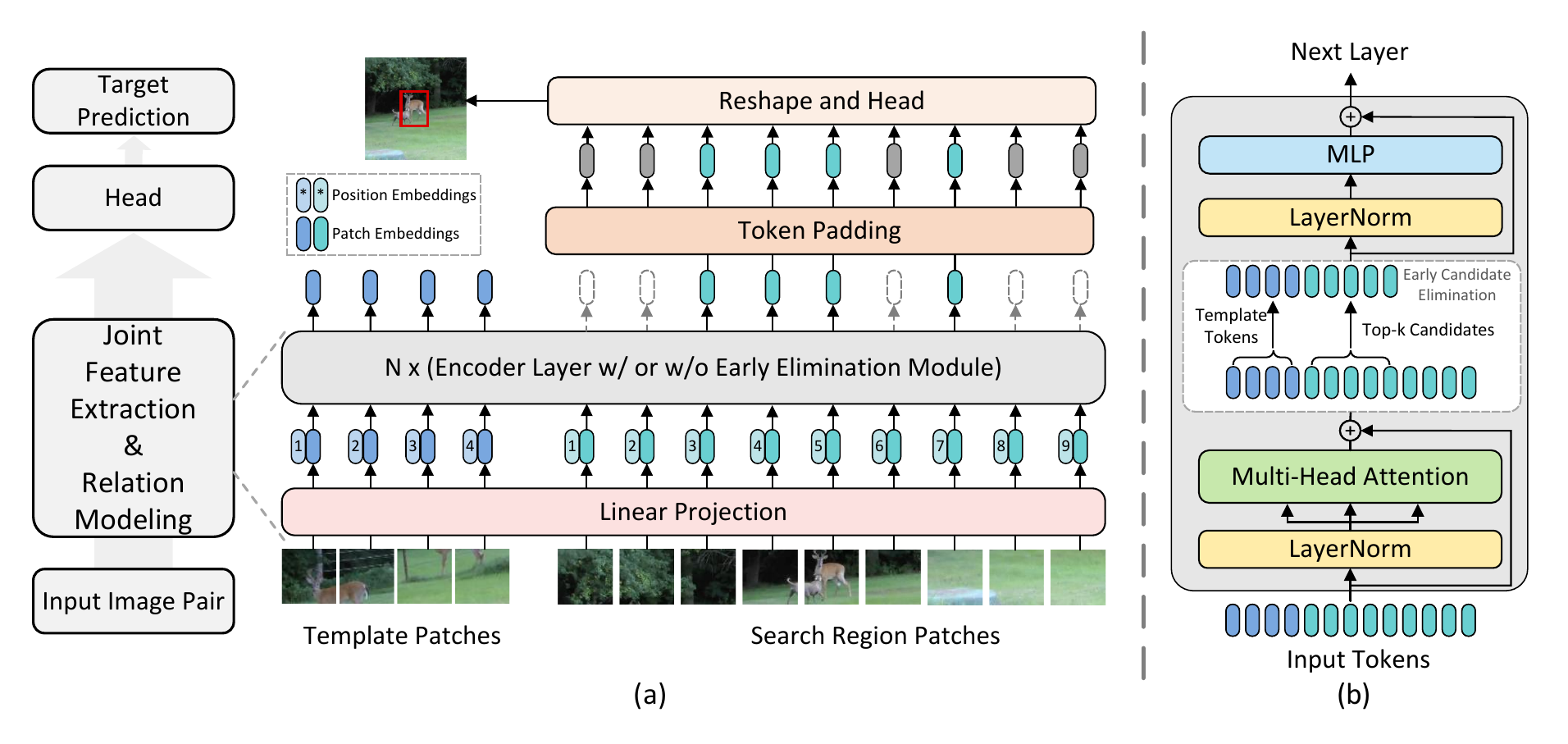}
\caption{\textbf{(a)} The overall framework of the proposed one-stream framework. The template and search region are split, flattened, and linear projected. Image embeddings are then concatenated and fed into Transformer encoder layers for joint feature extraction and relation modeling. \textbf{(b)} The structure of the encoder layer with early candidate elimination module, which is insert after the multi-head attention operation~\cite{transformer}.}
\label{fig:arch}
\end{figure}

This section describes the proposed one-stream tracker (OSTrack). The input image pairs are fed into a ViT backbone for simultaneous feature extraction and relation modeling, and the resulting search region features are directly adopted for subsequent target classification and regression. An overview of the model is shown in Fig.~\ref{fig:arch}(a).

\subsection{Joint Feature Extraction and Relation Modeling}
We propose to combine the feature extraction and relation modeling modules and construct a free information  flow between the contents of the template and the search region. The global contextual modeling capacity of self-attention~\cite{transformer} operation perfectly fits our goal, therefore, vanilla ViT~\cite{vit} is selected as the main body of OSTrack. Adopting the existing Vision Transformer architecture also provides a bunch of publicly available pre-trained models~\cite{mae, vit_in22k}, freeing us from the time-consuming pre-training stage.
The input of OSTrack is a pair of images, namely, the template image patch $z \in \mathbb{R}^{3\times H_z \times W_z}$ and the search region patch $x \in \mathbb{R}^{3\times H_x \times W_x}$. They are first split and flattened into sequences of patches $z_p \in \mathbb{R}^{N_z \times \left( 3 \cdot P^2 \right)}$ and $x_p \in \mathbb{R}^{N_x \times \left( 3 \cdot P^2 \right)}$, where $P \times P$ is the resolution of each patch, and $N_z = H_{z}W_{z}/P^2$, $N_x = H_{x}W_{x}/P^2$ are the number of patches of template and search region respectively. After that, a trainable linear projection layer with parameter $\boldsymbol{E}$ is used to project $z_p$ and $x_p$ into $D$ dimension latent space as in Eq.~\ref{eq:z_proj} and Eq.~\ref{eq:x_proj}, and the output of this projection is commonly called patch embeddings~\cite{vit}. Learnable 1D position embeddings $\boldsymbol{P}_z$ and $\boldsymbol{P}_x$ are added to the patch embeddings of the template and search region separately to produce the final template token embeddings $\boldsymbol{H}_{z}^{0} \in \mathbb{R}^{N_z \times D}$ and search region token embeddings $\boldsymbol{H}_{x}^{0} \in \mathbb{R}^{N_x \times D }$. 
\begin{align}
\label{eq:z_proj}
\boldsymbol{H}_{z}^{0} &= \left[\boldsymbol{z}_{p}^{1} \boldsymbol{E} ; \boldsymbol{z}_{p}^{2} \boldsymbol{E} ; \cdots ; \boldsymbol{z}_{p}^{N_z} \boldsymbol{E}\right]+\boldsymbol{P}_{z}, & & \boldsymbol{E} \in \mathbb{R}^{\left(3 \cdot P^{2}\right) \times D}, \boldsymbol{P}_{z} \in \mathbb{R}^{N_z \times D} \\
\label{eq:x_proj}
\boldsymbol{H}_{x}^{0} &= \left[\boldsymbol{x}_{p}^{1} \boldsymbol{E} ; \boldsymbol{x}_{p}^{2} \boldsymbol{E} ; \cdots ; \boldsymbol{x}_{p}^{N_z} \boldsymbol{E}\right]+\boldsymbol{P}_{x}, & & \boldsymbol{P}_{x} \in \mathbb{R}^{N_x \times D}
\end{align}
To verify whether adding addition identity embeddings (to indicate a token belonging to the template or search region as in BERT~\cite{bert}) or adopting relative positional embeddings are beneficial to the performance, we also conduct ablation studies and observe no significant improvement, thus they are omitted for simplicity (details can be found in the supplementary  material).

Token sequences $\boldsymbol{H}_{z}^{0}$ and $\boldsymbol{H}_{x}^{0}$ are then concatenated as $\boldsymbol{H}_{zx}^{0}  = [\boldsymbol{H}_{z}^{0} ; \boldsymbol{H}_{x}^{0}]$, and the resulting vector $\boldsymbol{H}_{zx}^{0}$ is then fed into several Transformer encoder layers~\cite{vit}. 
Unlike the vanilla ViT~\cite{vit}, we insert the proposed early candidate eliminating module into some of encoder layers as shown in Fig.~\ref{fig:arch}(b) for inference efficiency, and the technical details are presented in Sec.~\ref{subsec:token_drop}.
Notably, adopting the self-attention of concatenated features makes the whole framework highly parallelized compared to the cross-attention~\cite{transt}. Although template images are also fed into the ViT for each search frame, the impact on the inference speed is minor due to the highly parallel structure and the fact that the number of template tokens is small compared to the number of search region tokens.

\textbf{Analysis.}
From the perspective of the self-attention mechanism~\cite{transformer}, we further analyze the intrinsic reasons why the proposed framework is able to realize simultaneous feature extraction and relation modeling. The output of self-attention operation $\boldsymbol{A}$ in our approach can be written as:
\begin{equation}
\label{eq:attn}
\boldsymbol{A}=\operatorname{Softmax}\left(\frac{\boldsymbol{Q} \boldsymbol{K}^{\top}}{\sqrt{d_{k}}}\right) \cdot \boldsymbol{V} = \operatorname{Softmax}\left(\frac{[\boldsymbol{Q}_{z} ; \boldsymbol{Q}_{x}] [\boldsymbol{K}_{z} ; \boldsymbol{K}_{x}]^{\top}}{\sqrt{d_{k}}}\right) \cdot [\boldsymbol{V}_{z} ; \boldsymbol{V}_{x}], 
\end{equation}
where $\boldsymbol{Q}$, $\boldsymbol{K}$, and $\boldsymbol{V}$ are query, key and value matrices separately. The subscripts $z$ and $x$ denote matrix items belonging to the template and search region. The calculation of attention weights in Eq.~\ref{eq:attn} can be expanded to:
\begin{equation}
\label{eq:attn_expand}
\begin{aligned}
\operatorname{Softmax}\left(\frac{[\boldsymbol{Q}_{z} ; \boldsymbol{Q}_{x}] [\boldsymbol{K}_{z} ; \boldsymbol{K}_{x}]^{\top}}{\sqrt{d_{k}}}\right) &= 
\operatorname{Softmax}\left(\frac{[\boldsymbol{Q}_{z} \boldsymbol{K}_{z}^{\top},  \boldsymbol{Q}_{z} \boldsymbol{K}_{x}^{\top}; \boldsymbol{Q}_{x} \boldsymbol{K}_{z}^{\top}, \boldsymbol{Q}_{x} \boldsymbol{K}_{x}^{\top}]}{\sqrt{d_{k}}}\right) \\
&\triangleq [\boldsymbol{W}_{zz}, \boldsymbol{W}_{zx}; \boldsymbol{W}_{xz}, \boldsymbol{W}_{xx}],
\end{aligned}
\end{equation}
where $\boldsymbol{W_{zx}}$ is a measure of similarity between the template and the search region, and the rest are similar. The output $\boldsymbol{A}$ can be further written as:
\begin{equation}
\label{eq:value_expand}
\boldsymbol{A} = [\boldsymbol{W}_{zz} \boldsymbol{V}_{z} + \boldsymbol{W}_{zx} \boldsymbol{V}_{x} ; 
\boldsymbol{W}_{xz} \boldsymbol{V}_{z} + \boldsymbol{W}_{xx} \boldsymbol{V}_{x}].
\end{equation}
In the right part of Eq.~\ref{eq:value_expand}, $\boldsymbol{W}_{xz} \boldsymbol{V}_{z}$ is responsible for aggregating the iter-image feature (relation modeling) and $\boldsymbol{W}_{xx} \boldsymbol{V}_{x}$ aggregating the intra-image feature (feature extraction) based on the similarity of different image parts. Therefore, the feature extraction and relation modeling can be done with a self-attention operation.
Moreover, Eq.~\ref{eq:value_expand} also constructs a bi-direction information flow that allows mutual guidance of target-oriented feature extraction through the similarity learning.

\textbf{Comparisons with Two-Stream Transformer Fusion Trackers.} 
1) Previous two-stream Transformer fusion trackers~\cite{transt, swintrack} all adopt a Siamese framework, where the features of the template and search region are separately extracted first, and the Transformer layer is only adopted to fuse the extracted features. Therefore, the extracted features of these methods are not adaptive and may lose some discriminative information, which is irreparable.
In contrast, OSTrack directly concatenates linearly projected template and search region images at the first stage, so feature extraction and relation modeling are seamlessly integrated and target-oriented features can be extracted through the mutual guidance of the template and the search region.
2) Previous Transformer fusion trackers only employ ImageNet~\cite{imagenet} pre-trained backbone networks~\cite{resnet, swint} and leave Transformer layers randomly initialized, which degrades the convergence speed, while OSTrack benefits from pre-trained ViT models for faster convergence.
3) The one-stream framework provides the possibility of identifying and discarding useless background regions for further improving the model performance and inference speed as presented in Sec.~\ref{subsec:token_drop}.

\begin{figure}[t]
\centering
\setlength{\abovecaptionskip}{0.05cm}
\includegraphics[width=0.65\columnwidth, keepaspectratio]{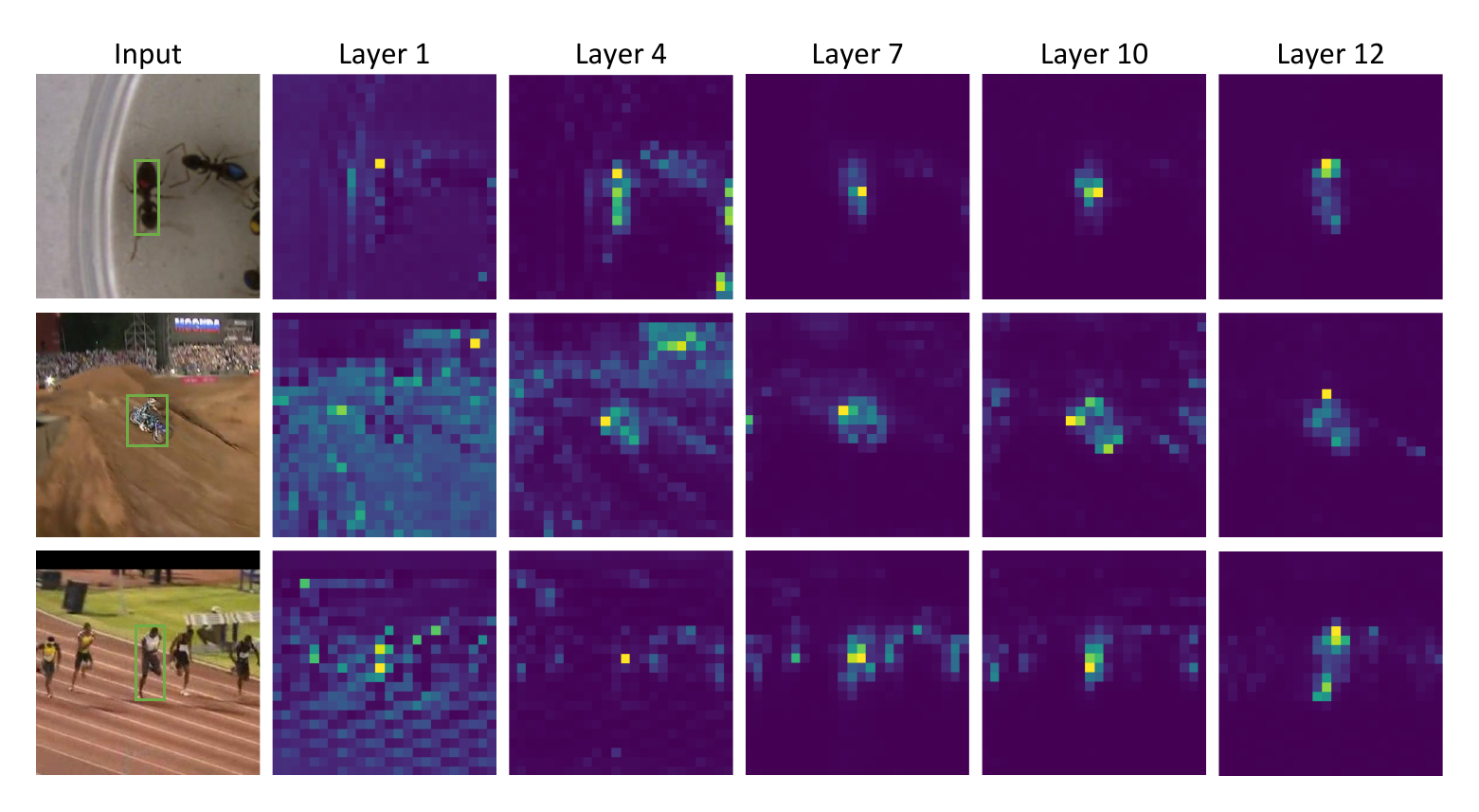}
\caption{Visualization of the attention weights of search region corresponding to the center part of template after different ViT layers, the \textcolor{green}{green} rectangles indicate target objects. It can be seen as an estimate of the similarity between the target and each position of the search region.}
\label{fig:vis_attn}
\end{figure}

\subsection{Early Candidate Elimination}
\label{subsec:token_drop}
Each token of the search region can be regarded as a target candidate and each template token can be considered as a part of the target object. Previous trackers keep all candidates during feature extraction and relation modeling, while background regions are not identified until the final output of the network (\ie, classification score map). However, our one-stream framework provides a strong prior on the similarity between the target and each candidate. As shown in Fig.~\ref{fig:vis_attn}, the attention weights of the search region highlight the foreground objects in the early stage of ViT (\eg, layer 4), and then progressively focus on the target. This property makes it possible to progressively identify and eliminate candidates belonging to the background regions inside the network. Therefore, we propose an early candidate elimination module that progressively eliminates candidates belonging to the background in the early stages of ViT to lighten the computational burden and avoid the negative impact of noisy background regions on feature learning.

\textbf{Candidate Elimination.} Recall that the self-attention operation in ViT can be seen as a spatial aggregation of tokens with normalized importances~\cite{transformer}, which is measured by the dot product similarity between each token pair. Specifically, each template token $\boldsymbol{h}_z^i, 1 \leq i \leq N_z$ is calculated as:
\begin{equation}
\boldsymbol{h}_z^i =\operatorname{Softmax}\left(\frac{\boldsymbol{q}_{i} \cdot [\boldsymbol{K}_{z} ; \boldsymbol{K}_{x}]^{\top}}{\sqrt{d}}\right) \cdot \boldsymbol{V}=[\boldsymbol{w}_{z}^{i} ; \boldsymbol{w}_{x}^{i}] \cdot \boldsymbol{V} ,
\end{equation}
where $\boldsymbol{q}_{i}$, $\boldsymbol{K}_{z}$, $\boldsymbol{K}_{x}$ and $\boldsymbol{V}$ denote the query vector of token $\boldsymbol{h}_z^i$, the key matrix corresponding to the template, the key matrix corresponding to the search region and the value matrix. The attention weight $\boldsymbol{w}_{x}^{i}$ determines the similarity between the template part $\boldsymbol{h}_z^i$ and all search region tokens (candidates). The $j-th$ item ($1 \leq j \leq n$, $n$ is the number of input search region tokens) of $\boldsymbol{w}_{x}^{i}$ determines the similarity between $\boldsymbol{h}_z^i$ and the $j-th$ candidate. However, the input templates usually include background regions that introduce noise when calculating the similarity between the target and each candidate. Therefore, instead of summing up the similarity 
of each candidate to all template parts $\boldsymbol{w}_{x}^{i}$, $i=1,\ldots, N_z$, we take $\boldsymbol{w}_{x}^{\phi}, \phi = \lfloor \frac{W_z}{2} \rfloor + W_z \cdot \lfloor \frac{H_z}{2} \rfloor $ ($\phi-th$ token corresponding to the center part of the original template image) as the representative similarity. This is fairly reasonable as the center template part has aggregated enough information through self-attention to represent the target. In the supplementary, we compare the effect of different template token choices.
Considering that multi-head self-attention is used in ViT, there are multiple similarity scores $\boldsymbol{w}_{x}^{\phi}(m)$, where $m = 1, . . . ,M$ and $M$ is the total number of attention heads~\cite{transformer}. We average the similarity scores of all heads by $\overline{\boldsymbol{w}}_{x}^{\phi} = \sum_{m=1}^{M} \boldsymbol{w}_{x}^{\phi}(m) / M$, which serves as the final similarity score of the target and each candidate. One candidate is more likely to be a background region if its similarity score with the target is relatively small. Therefore, we only keep the candidates corresponding to the $k$ largest (top-$k$) elements in $\overline{\boldsymbol{w}}_{x}^{\phi}$ ($k$ is a hyperparameter, and we define the token keeping ratio as $\rho = k / n$), while the remaining candidates are eliminated. The proposed candidate elimination module is inserted after the multi-head attention operation~\cite{transformer} in the encoder layer, which is illustrated in Fig.~\ref{fig:arch}(b). In addition, the original order of all remaining candidates is recorded so that it can be recovered in the final stage.

\textbf{Candidate Restoration.} The aforementioned candidate elimination module disrupts the original order of the candidates, making it impossible to reshape the candidate sequence back into the feature map as described in Sec.~\ref{subsec:head}, so we restore the original order of the remaining candidates and then pad the missing positions. Since the discarded candidates belong to the irrelevant background regions, they will not affect the classification and regression tasks. In other words, they just act as placeholders for the reshaping operation. Therefore, we first restore the order of the remaining candidates and then zero-pad in between them.

\begin{figure}[t]
\centering
\setlength{\abovecaptionskip}{0.05cm}
\includegraphics[width=\columnwidth, keepaspectratio]{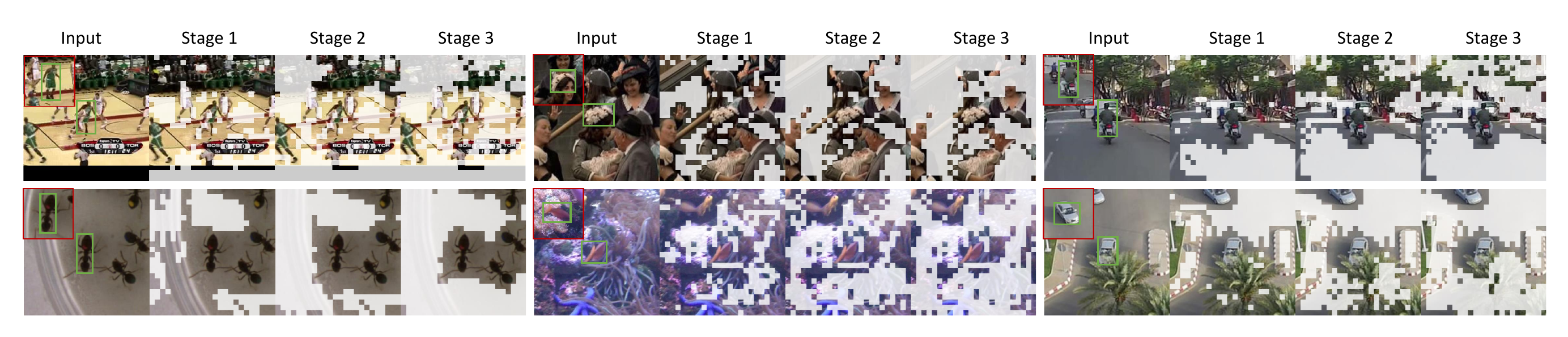}
\caption{Visualization of the progressive early candidate elimination process. The main body of ``Input'' is the search region image, and the upper left corner shows the corresponding template image. The \textcolor{green}{Green} rectangles indicate target objects and the masked regions represent the discarded candidates. The results show that our method can gradually identify and discard the candidates belonging to the background regions.}
\label{fig:vis_drop}
\end{figure}

\textbf{Visualization.} To further investigate the behavior of the early candidate elimination module, we visualize the progressive process in Fig.~\ref{fig:vis_drop}. By iteratively discarding the irrelevant tokens in the search region, OSTrack not only largely lightens the computation burden but also avoids the negative impact of noisy background regions on feature learning. 

\subsection{Head and Loss}
\label{subsec:head}
We first re-interpret the padded sequence of search region tokens to a 2D spatial feature map and then feed it into a fully convolutional network (FCN), which consists of $L$ stacked Conv-BN-ReLU layers for each output. Outputs of the FCN contain the target classification score map $\boldsymbol{P} \in[0,1]^{\frac{H_x}{P} \times \frac{W_x}{P}} $, the local offset $\boldsymbol{O} \in[0,1)^{2 \times \frac{H_x}{P} \times \frac{W_x}{P}}$ to compensate the discretization error caused by reduced resolution and the normalized bounding box size (\ie width and height)  $\boldsymbol{S} \in[0,1]^{2 \times \frac{H_x}{P} \times \frac{W_x}{P}} $.
The position with highest classification score is considered to be target position, \ie, $(x_d, y_d) = \arg\max_{(x,y)} \boldsymbol{P}_{xy}$ and the finial target bounding box is obtained as:
\begin{align}
\label{equ:bbox}
    \hspace{-0.5mm} (x, y, w, h) = (& x_d + \boldsymbol{O}(0, x_d, y_d), y_d + \boldsymbol{O}(1, x_d, y_d), \boldsymbol{S}(0, x_d, y_d), \boldsymbol{S}(1, x_d, y_d)).
\end{align}

During training, both classification and regression losses are used. We adopt the weighted focal loss~\cite{cornernet} for classification (see the supplementary for more details). With the predicted bounding box, $\ell_1$ loss and the generalized IoU loss~\cite{GIoULoss} are employed for bounding box regression. Finally, the overall loss function is:

\begin{equation}
	\label{equ-loss-loc}
	\begin{aligned}
		L_{track}=L_{cls} + \lambda_{iou}L_{iou} + \lambda_{L_1}L_1,
	\end{aligned}
\end{equation}
where $\lambda_{iou}=2$ and $\lambda_{L_1}=5$ are the regularization parameters in our experiments as in~\cite{stark}.

\section{Experiments}
After introducing the implementation details, this section first presents a comparison of OSTrack with other state-of-the-art methods on seven different benchmarks. Then, ablation studies are provided to analyze the impact of each component and different design choices.

\subsection{Implementation Details}
Our trackers are implemented in Python using PyTorch. The models are trained on 4 NVIDIA A100 GPUs and the inference speed is tested on a single NVIDIA RTX2080Ti GPU.

\textbf{Model.} 
The vanilla ViT-Base~\cite{vit} model pre-trained with MAE~\cite{mae} is adopted as the backbone for joint feature extraction and relation modeling. The head is a lightweight FCN, consisting of 4 stacked Conv-BN-ReLU layers for each of three outputs. The keeping ratio $\rho$ of each candidate elimination module is set as 0.7, and a total of three candidate elimination modules are inserted at layers 4, 7, and 10 of ViT respectively, following~\cite{dynamicvit}.
We present two variants with different input image pair resolution for showing the scalability of OSTrack:
\begin{itemize}

	\item{\textbf{OSTrack-256.} Template: 128$\times$128 pixels; Search region: 256$\times$256 pixels.}
	
	\item{\textbf{OSTrack-384.} Template: 192$\times$192 pixels; Search region: 384$\times$384 pixels.}
    
\end{itemize}

\textbf{Training.} 
The training splits of COCO~\cite{coco}, LaSOT~\cite{lasot}, GOT-10k~\cite{got10k} (1k forbidden sequences from GOT-10k training set are removed following the convention~\cite{stark}) and TrackingNet~\cite{trackingnet} are used for training. Common data augmentations including horizontal flip and brightness jittering are used in training. Each GPU holds 32 image pairs, resulting in a total batch size of 128. We train the model with AdamW optimizer~\cite{adamw}, set the weight decay to $10^{-4}$, the initial learning rate for the backbone to $4\times10^{-5}$ and other parameters to $4\times10^{-4}$, respectively. The total training epochs are set to 300 with 60k image pairs per epoch and we decrease the learning rate by a factor of 10 after 240 epochs.

\textbf{Inference.} 
During inference, Hanning window penalty is adopted to utilize positional prior in tracking following the common practice~\cite{ocean, transt}. Specifically, we simply multiply the classification map $\boldsymbol{P}$ by the Hanning window with the same size, and the box with the highest score after multiplication will be selected as the tracking result.

\subsection{Comparison with State-of-the-arts}
To demonstrate the effectiveness of the proposed models, we compare them with state-of-the-art (SOTA) trackers on seven different benchmarks.
\begin{table}[t]
\caption[Caption for LOF]{Comparison with state-of-the-arts on four large-scale benchmarks: LaSOT, LaSOT$_{\text{ext}}$, TrackingNet and GOT-10k\protect\footnotemark . The best two results are shown in  \rankfirst{red} and \ranksecond{blue} fonts.}
\vspace{-3mm}

\label{tab:sota_compare1}
\begin{center}
\resizebox{0.95\linewidth}{!}{
\begin{tabular}{c|c|ccc|ccc|ccc|ccc}
\hline
\multirow{2}{*}{Method} &
  \multirow{2}{*}{Source} &
  \multicolumn{3}{c|}{LaSOT~\cite{lasot}} &
  \multicolumn{3}{c|}{LaSOT$_{\text{ext}}$~\cite{lasot_ext}} &
  \multicolumn{3}{c|}{TrackingNet~\cite{trackingnet}} &
  \multicolumn{3}{c}{GOT-10k$^*$ ~\cite{got10k}} \\ \cline{3-14} 
                                      &           & AUC  & P$_{Norm}$ & P    & AUC  & P$_{Norm}$ & P    & AUC  & P$_{Norm}$ & P    & AO   & SR$_{0.5}$ & SR$_{0.75}$ \\ \hline
SiamFC~\cite{siamfc}                  & ECCVW16 & 33.6 & 42.0       & 33.9 & 23.0 & 31.1       & 26.9 & 57.1 & 66.3       & 53.3 & 34.8 & 35.3       & 9.8         \\
MDNet~\cite{mdnet}                    & CVPR16  & 39.7 & 46.0       & 37.3 & 27.9 & 34.9       & 31.8 & 60.6 & 70.5       & 56.5 & 29.9 & 30.3       & 9.9         \\
ECO~\cite{eco}                        & ICCV17  & 32.4 & 33.8       & 30.1 & 22.0 & 25.2       & 24.0 & 55.4 & 61.8       & 49.2 & 31.6 & 30.9       & 11.1        \\
SiamPRN++~\cite{siamrpn++}            & CVPR19  & 49.6 & 56.9       & 49.1 & 34.0 & 41.6       & 39.6 & 73.3 & 80.0       & 69.4 & 51.7 & 61.6       & 32.5        \\
DiMP~\cite{dimp}                      & ICCV19  & 56.9 & 65.0       & 56.7 & 39.2 & 47.6       & 45.1 & 74.0 & 80.1       & 68.7 & 61.1 & 71.7       & 49.2        \\
SiamR-CNN~\cite{siamrcnn}             & CVPR20  & 64.8 & 72.2       & -    & -    & -          & -    & 81.2 & 85.4       & 80.0 & 64.9 & 72.8       & 59.7        \\
MAMLTrack~\cite{maml_track}           & CVPR20  & 52.3 & -          & -    & -    & -          & -    & 75.7 & 82.2       & 72.5 & - & - & - \\
LTMU~\cite{ltmu}                & CVPR20  & 57.2 & -          & 57.2    & 41.4    & 49.9          & 47.3    & - & -       & - & - & - & - \\
Ocean~\cite{ocean}                    & ECCV20  & 56.0 & 65.1       & 56.6 & -    & -          & -    & -    & -          & -    & 61.1 & 72.1       & 47.3        \\
TrDiMP~\cite{trdimp}                  & CVPR21  & 63.9 & -          & 61.4 & -    & -          & -    & 78.4 & 83.3       & 73.1 & 67.1 & 77.7       & 58.3        \\
TransT~\cite{transt}                  & CVPR21  & 64.9 & 73.8       & 69.0 & -    & -          & -    & 81.4 & 86.7       & 80.3 & 67.1 & 76.8       & 60.9        \\
AutoMatch~\cite{automatch}            & ICCV21  & 58.3 & -          & 59.9 & -    & -          & -    & 76.0 & -          & 72.6 & 65.2 & 76.6       & 54.3        \\
STARK~\cite{stark}                    & ICCV21  & 67.1 & 77.0       & -    & -    & -          & -    & 82.0 & 86.9       & -    & 68.8   & 78.1       & 64.1        \\
KeepTrack~\cite{keeptrack}            & ICCV21  & 67.1 & 77.2       & 70.2 & \ranksecond{48.2} & -          & -    & -    & -          & -    & -    & -          & -           \\
SwinTrack-B~\cite{swintrack}          & arXiv21  & \ranksecond{69.6} & 78.6       & 74.1 & 47.6 & \ranksecond{58.2}       & \ranksecond{54.1}   & 82.5   &  87.0          & 80.4    & 69.4    & 78.0          & 64.3  \\ \hline

OSTrack-256                        & Ours    & 69.1    & \ranksecond{78.7}          & \ranksecond{75.2}    & 47.4    & 57.3         & 53.3    & \ranksecond{83.1}    & \ranksecond{87.8}          & \ranksecond{82.0}   & \ranksecond{71.0}    & \ranksecond{80.4}          & \ranksecond{68.2}          \\
OSTrack-384 &
  Ours &
  \rankfirst{71.1} &
  \rankfirst{81.1} &
  \rankfirst{77.6} &
  \rankfirst{50.5} &
  \rankfirst{61.3} &
  \rankfirst{57.6} &
  \rankfirst{83.9} &
  \rankfirst{88.5} &
  \rankfirst{83.2} &
  \rankfirst{73.7} &
  \rankfirst{83.2} &
  \rankfirst{70.8} \\ \hline
 \end{tabular}}
\end{center}


\end{table}
\begin{table}[t]
\centering
\caption{Comparison with state-of-the-arts on three benchmarks: NFS~\cite{nfs}, UAV123~\cite{uav} and TNL2K~\cite{tnl2k}. AUC(\%) scores are reported. The best two results are shown in  \rankfirst{red} and \ranksecond{blue} fonts.}

\resizebox{0.95\linewidth}{!}{
\begin{tabular}{c|ccccccccc|cc}
\hline
 &  \begin{tabular}[c]{@{}c@{}}SiamFC\\ \cite{siamfc}\end{tabular}
 & \begin{tabular}[c]{@{}c@{}}RT-MDNet\\ \cite{rtrmdnet}\end{tabular}
 & \begin{tabular}[c]{@{}c@{}}ECO\\ \cite{eco}\end{tabular}
 & \begin{tabular}[c]{@{}c@{}}Ocean\\ \cite{ocean}\end{tabular}
 & \begin{tabular}[c]{@{}c@{}}ATOM\\ \cite{atom}\end{tabular}
 & \begin{tabular}[c]{@{}c@{}}DiMP50\\ \cite{dimp}\end{tabular}
 & \begin{tabular}[c]{@{}c@{}}STMTrack\\ \cite{stmtrack}\end{tabular}
 & \begin{tabular}[c]{@{}c@{}}TransT\\ \cite{transt}\end{tabular}
 & \begin{tabular}[c]{@{}c@{}}STARK\\ \cite{stark}\end{tabular}
 & \begin{tabular}[c]{@{}c@{}}OSTrack\\ -256\end{tabular} & \begin{tabular}[c]{@{}c@{}}OSTrack\\ -384\end{tabular} \\ \hline
NFS    & 37.7 & 43.3 & 52.2 & 49.4  & 58.3 & 61.8 & -    & 65.3 & \ranksecond{66.2}  & 64.7 & \rankfirst{66.5} \\
UAV123 & 46.8 & 52.8 & 53.5 & 57.4  & 63.2 & 64.3 & 64.7 & 68.1 & 68.2 & \ranksecond{68.3} & \rankfirst{70.7} \\ 
TNL2K & 29.5 & -   & 32.6 &  38.4 & 40.1 & 44.7 & - & 50.7 & -  & \ranksecond{54.3} & \rankfirst{55.9} \\ \hline

\end{tabular}}

\label{tab:sota_compare2}
\end{table}

{\textbf{GOT-10k.}} GOT-10k~\cite{got10k} test set employs the one-shot tracking rule, \ie, it requires the trackers to be trained only on the GOT-10k training split, and the object classes between train and test splits are not overlapped. We follow this protocol to train our model and evaluate the results by submitting them to the official evaluation server. As reported in Tab.~\ref{tab:sota_compare1}, OSTrack-384 and OSTrack-256 outperform SwinTrack-B~\cite{swintrack} by 1.6\% and 4.3\% in AO. The SR$_{0.75}$ score of OSTrack-384 reaches 70.8\%, outperforming SwinTrack-B by 6.5\%, which verifies the capability of our trackers in both accurate target-background discrimination and bounding box regression. Moreover, the high performance on this one-shot tracking benchmark demonstrates that our one-stream tracking framework can extract more discriminative features for unseen classes by mutual guidance\footnotetext{We add the symbol $*$ to GOT-10k if the corresponding models are trained following the one-shot protocol, otherwise they are trained with all training data.}.

{\textbf{LaSOT.}} LaSOT~\cite{lasot} is a challenging large-scale long-term tracking benchmark, which contains 280 videos for testing. We compare the result of the OSTrack with previous SOTA trackers in Tab.~\ref{tab:sota_compare1}. The results show that the proposed tracker with smaller input resolution, \ie, OSTrack-256, already obtains comparable performance with SwinTrack-B~\cite{swintrack}. Besides, OSTrack-256 runs at a fast inference speed of 105.4 FPS, being 2x faster than SwinTrack-B (52 FPS), which indicates that OSTrack achieves an excellent balance between accuracy and inference speed. By increasing the input resolution, OSTrack-384 further improves the AUC on LaSOT to 71.1\% and sets a new state-of-the-art.

{\textbf{TrackingNet.}} The TrackingNet~\cite{trackingnet} benchmark contains 511 sequences for testing, which covers diverse target classes. Tab.~\ref{tab:sota_compare1} shows that  OSTrack-256 and OSTrack-384 surpass SwinTrack-B~\cite{swintrack} by 0.6\% and 1.4\% in AUC separately. Moreover, both models are faster than SwinTrack-B.

{\textbf{LaSOT$_{\text{ext}}$.}} LaSOT$_{\text{ext}}$~\cite{lasot_ext} is a recently released extension of LaSOT, which consists of 150 extra videos from 15 object classes. Tab~\ref{tab:sota_compare1} presents the results. Previous SOTA tracker KeepTrack~\cite{keeptrack} designs a complex association network and runs at 18.3 FPS. In contrast, our simple one-stream tracker OSTrack-256 shows slightly lower performance but runs at 105.4 FPS. OSTrack-384 sets a new state-of-the-art AUC score of 50.5\% while runs in 58.1 FPS, which is 2.3\% higher in AUC score and 3x faster in speed.

{\textbf{NFS, UAV123 and TNL2K.}} We also evaluate our tracker on three additional benchmarks: NFS~\cite{nfs}, UAV123~\cite{uav} and TNL2K~\cite{tnl2k} includes 100, 123, and 700 video sequences, separately. The results in Tab.~\ref{tab:sota_compare2} show that OSTrack-384 achieves the best performance on all three benchmarks, demonstrating the strong generalizability of OSTrack.

\subsection{Ablation Study and Analysis}
\label{subsec:ablation}
\textbf{The Effect of Early Candidate Elimination Module}
Tab.~\ref{tab:sota_compare1} shows that increasing the input resolution of the input image pairs can bring significant performance gain. However, the quadratic complexity with respect to the input resolution makes simply increasing the input resolution unaffordable in inference time. The proposed early candidate elimination module addresses the above problem well. We present the effect of the early candidate elimination module from the aspects of inference speed (FPS), multiply-accumulate computations (MACs), and tracking performance on multiple benchmarks in Tab.~\ref{tab:token_drop_effect}. The effect on different input search region resolutions is also presented. Tab.~\ref{tab:token_drop_effect} shows that the early candidate elimination module can significantly decrease the calculation and increase the inference speed, while slightly boosting the performance in most cases. This demonstrates that the proposed module alleviates the negative impact brought by the noisy background regions on feature learning. For example, adding the early candidate elimination module in OSTrack-256 decreases the MACs by 25.9\% and increases the tracking speed by 13.2\%, and the LaSOT AUC is increased by 0.4\%. Furthermore, larger input resolution benefits more from this module, \eg, OSTrack-384 shows a 40.3\% increase in speed.

\begin{table}[t]
\centering
\caption{The effect of our proposed early candidate elimination module on the inference speed, MACs and tracking performance on LaSOT, GOT-10k and TrackingNet benchmarks, and w/o and w/ denote the models with or without early candidate elimination module separately.}
\label{tab:token_drop_effect}
\resizebox{\linewidth}{!}{
\begin{tabular}{c|cc|cc|cc|cc|cc}
\hline
\multirow{2}{*}{\begin{tabular}[c]{@{}c}Input\\ Resolution\end{tabular} } &
  \multicolumn{2}{c|}{FPS} &
  \multicolumn{2}{c|}{MACs (G)} &
  \multicolumn{2}{c|}{LaSOT AUC (\%)} &
  \multicolumn{2}{c|}{TrackingNet AUC (\%)} &
  \multicolumn{2}{c}{GOT-10k$^*$ AO (\%)} \\ \cline{2-11} 
    & w/o  & w/             & w/o  & w/            & w/o  & w/          & w/o & w/ & w/o & w/ \\ \hline
256x256 & 93.1 & 105.4(\textcolor{blue}{+13.2\%}) & 29.0   & 21.5(\textcolor{blue}{-25.9\%}) & 68.7 & 69.1(\textcolor{blue}{+0.4})  &   82.9  & 83.1(\textcolor{blue}{+0.2})  &  71.0   &  71.0(\textcolor{blue}{+0.0})  \\
384x384 & 41.4 & 58.1(\textcolor{blue}{+40.3\%})  & 65.3 & 48.3(\textcolor{blue}{-26.0\%}) & 71.0   & 71.1(\textcolor{blue}{+0.1})  &  83.5   &  83.9(\textcolor{blue}{+0.4}) &   73.5  &  73.7(\textcolor{blue}{+0.2}) \\ \hline
\end{tabular}}
\end{table}

\begin{table}[t]
\centering
\caption{The effect of different pre-training methods. All the models are trained without the early candidate elimination module.}
\label{tab:pretrain}
\resizebox{0.7\linewidth}{!}{
\begin{tabular}{c|ccc|ccc|ccc}
\hline
\multirow{2}{*}{Trackers} & \multicolumn{3}{c|}{LaSOT} & \multicolumn{3}{c|}{TrackingNet} & \multicolumn{3}{c}{GOT-10k} \\ \cline{2-10} 
             & Success & P$_{Norm}$ & P    & Success & P$_{Norm}$ & P    & AO    & SR$_{0.5}$ & SR$_{0.75}$ \\ \hline
No pre-training & 60.4    & 70.0   & 62.8 & 77.5    & 83.0   & 73.8 & 62.7 & 72.8   & 53.7   \\
ImageNet-1k  & 66.1    & 75.8   & 70.6 & 82.0    & 86.7   & 80.1 & 69.7 & 79.0   & 65.6   \\
ImageNet-21k & 66.9    & 76.3   & 71.2 & 82.4    & 86.9   & 80.1 & 70.2 & 80.7   & 65.4   \\
MAE         & \best{68.7}    & \best{78.1}   & \best{74.6} & \best{82.9}    & \best{87.5}   & \best{81.6} & \best{73.6} & \best{83.0}   & \best{71.7}   \\ \hline
\end{tabular}}


\end{table}

\textbf{Different Pre-training Methods.}
While previous Transformer fusion trackers~\cite{transt, stark, swintrack} random initialize the weights of Transformer layers, our joint feature learning and relation modeling module can directly benefit from the pre-trained weights. We further investigate the effect of different pre-training methods on the tracking performance by comparing four different pre-training strategies: no pre-training; ImageNet-1k~\cite{imagenet} pre-trained model provided by~\cite{deit}; ImageNet-21k~\cite{imagenet21k} pre-trained model provided by~\cite{vit_in22k}; unsupervised pre-training model MAE~\cite{mae}. As the results in Tab.~\ref{tab:pretrain} show, pre-training is necessary for the model weights initialization. Interestingly, we also observe that the unsupervised pre-training method MAE brings better tracking performance than the supervised pre-training ones using ImageNet. We hope this can inspire the community for designing better pre-training strategies tailored for the tracking task.

\textbf{Aligned Comparison with SOTA Two-stream Trackers.}
One may wonder whether the performance gain is brought by the proposed one-stream structure or purely by the superiority of ViT. We thus compare our method with two SOTA two-stream Transformer fusion trackers~\cite{stark, swintrack} by eliminating the influencing factors of backbone and head structure. To be specific, we align two previous SOTA two-stream trackers (STRAK-S~\cite{stark} and SwinTrack~\cite{swintrack}) with ours for fair comparison as follows: replacing their backbones with the same pre-trained ViT and setting the same input resolution, head structure, and training objective as OSTrack-256. The remaining experimental settings are kept the same as in the original paper.
As shown in Tab.~\ref{tab:align_compare}, our re-implemented two-stream trackers show comparable or stronger performance compared to the initially published performance, but still lag behind OSTrack, which demonstrates the effectiveness of our one-stream structure. We also observe that OSTrack significantly outperforms the previous two-stream trackers on the one-shot benchmark GOT-10k, which further proves the advantage of our one-stream framework in the challenging scenario. Actually, the discriminative power of features extracted by the two-stream framework is limited since the object classes in the testing set are completely different from the training set. Whereas, by iterative interaction between the features of the template and search region, OSTrack can extract more discriminative features through mutual guidance.
Different from the two-stream SOTA trackers, OSTrack neglects the extra heavy relation modeling module while still keeping the high parallelism of joint feature extraction and relation modeling module. Therefore, when the same backbone network is adopted, the proposed one-stream framework is much faster than STARK (40.2 FPS faster) and SwinTrack (25.6 FPS faster). Besides, OSTrack requires fewer training image pairs to converge.

\begin{figure}[t]
\centering
\setlength{\abovecaptionskip}{0.05cm}
\includegraphics[width=0.8\columnwidth, keepaspectratio]{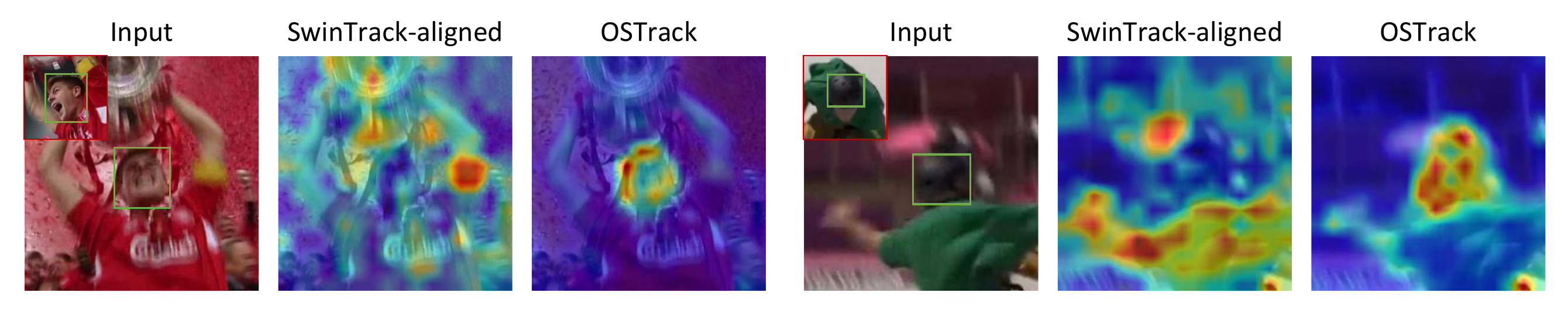}
\caption{Visualization of discriminative regions (\ie, activation maps) of backbone features extracted by OSTrack and two-stream tracker (SwinTrack-aligned).}
\label{fig:motivation}
\end{figure}

\textbf{Discriminative Region Visualization.} 
To better illustrate the effectiveness of the proposed one-stream tracker, we visualize the discriminative regions of the backbone features extracted by OSTrack and a SOTA two-stream tracker (SwinTrack-aligned) in Fig.~\ref{fig:motivation}. As can be observed, due to the lack of target awareness, features extracted by the backbone of SwinTrack-aligned show limited target-background discriminative power and may lose some important target information (\eg, head and helmet in Fig.~\ref{fig:motivation}), which is irreparable. In contrast, OSTrack can extract discriminative target-oriented features, since the proposed early fusion mechanism enables relation modeling between the template and search region at the first stage. 

\begin{table}[t!]
\centering
\caption{Comparison with re-implemented previous SOTA trackers aligned with OSTrack. Here OSTrack is trained without the early candidate elimination module for fair comparison and ``aligned'' denotes that the backbone, head, loss and input resolution are kept the same as OSTrack.}
\label{tab:align_compare}
\resizebox{0.85\linewidth}{!}{
\begin{tabular}{c|ccc|ccc|ccc|c|c}
\hline
\multirow{2}{*}{Trackers} &
  \multicolumn{3}{c|}{LaSOT} &
  \multicolumn{3}{c|}{TrackingNet} &
  \multicolumn{3}{c|}{GOT-10k$^*$} &
  \multirow{2}{*}{FPS} &
  \multirow{2}{*}{Traing Pairs($\times 10^6$)} \\ \cline{2-10}
       & Success & P$_{Norm}$ & P    & Success & P$_{Norm}$ & P    & AO    & SR$_{0.5}$ & SR$_{0.75}$ &      &    \\ \hline
STARK-aligned  & 67.6    & 76.3   & 72.8 & 82.6    & 87.4   & 81.5 & 68.8 & 78.4   & 65.6   & 52.9 & 30 \\
SwinTrack-aligned  & 68.0    & 77.6   & 73.9 & \best{82.9}    & \best{87.6}   & \best{81.6} & 69.5 & 79.2   & 65.0   & 67.5 & 39.3 \\
OSTrack   & \best{68.7}    & \best{78.1}   & \best{74.6} & \best{82.9}    & 87.5   & \best{81.6} & \best{71.0} & \best{80.3}   & \best{68.2}   & \best{93.1} & \best{18}  \\ \hline
\end{tabular}}


\end{table}

\section{Conclusion}

This work proposes a simple, neat, and high-performance one-stream tracking framework based on Vision Transformer, which breaks out of the Siamese-like pipeline. The proposed tracker combines the feature extraction and relation modeling tasks, and shows a good balance between performance and inference speed.
In addition, we further propose an early candidate elimination module that progressively discards search region tokens belonging to background regions, which significantly boosts the tracking inference speed.
Extensive experiments show that the proposed one-stream trackers perform much better than previous methods on multiple benchmarks, especially under the one-shot protocol. We expect this work can attract more attention to the one-stream tracking framework. \\

\noindent\textbf{Acknowledgments.} This work is partially supported by Natural Science Foundation
of China (NSFC): 61976203 and 61876171. Thanks Zhipeng Zhang for his helpful suggestions.

\clearpage
%
%
\bibliographystyle{splncs04}
\bibliography{egbib}

\clearpage
\section*{Appendix}
\beginappendix
\section{More Implementation Details}
\label{sec:more_imple}
\textbf{Training Details.}
In OSTrack-256, the input sizes of templates and search regions are $128 \times 128$ pixels and $256 \times 256$ pixels respectively, corresponding to $2^2$ and $4^2$ times of the target bounding box area. In OSTrack-384, the input sizes of templates and search regions are $192 \times 192$ pixels and $384 \times 384$ pixels, corresponding to $2^2$ and $5^2$ times of the target bounding box area.
For the GOT-10k test benchmark~\cite{got10k}, which requires training the models with only the training split of GOT-10k  (one-shot setting), we set the total training epoch to 100 with 60k image pairs per epoch, and we decrease the learning rate by a factor of 10 after 80 epochs. The other settings are kept consistent with the models trained with all datasets.

\textbf{Classification Loss.} 
We adopt the weighted focal loss~\cite{cornernet} for classification. Specifically, for each ground truth target center  $\hat{p}$ and its corresponding low-resolution equivalent  $\tilde{p}=[\tilde{p}_x, \tilde{p}_y]$, the ground truth heatmap can be generated using a Gaussian kernel as $\hat{\boldsymbol{P}}_{xy}=\exp \left(-\frac{\left(x-\tilde{p}_{x}\right)^{2}+\left(y-\tilde{p}_{y}\right)^{2}}{2 \sigma_{p}^{2}}\right)$, where $\sigma$ is an object size-adaptive standard deviation~\cite{cornernet}. The Gaussian weighted focal loss can be formulated as:
\begin{equation}
\setlength{\belowdisplayskip}{3pt} 
\setlength{\belowdisplayshortskip}{3pt}
\setlength{\abovedisplayskip}{3pt}
\setlength{\abovedisplayshortskip}{3pt}
    L_{cls} = - \sum_{xy}
    \begin{cases}
        (1 - \boldsymbol{P}_{xy})^{\alpha} 
        \mathrm{log}(\boldsymbol{P}_{xy}), & \!\text{if}\ \hat{\boldsymbol{P}}_{xy} = 1\vspace{2mm}\\
        (1 - \hat{\boldsymbol{P}}_{xy})^{\beta} (\boldsymbol{P}_{xy})^{\alpha} \mathrm{log}(1 - \boldsymbol{P}_{xy}), & \!\text{otherwise}
    \end{cases}
\end{equation}
where $\alpha$ and $\beta$ are hyper-parameters and we set $\alpha = 2$ and $\beta = 4$ as in ~\cite{cornernet, centernet}.

\textbf{Position Embeddings.}
The length of the position embeddings in the pre-trained ViT is different from the length of the input template and search region embeddings. Therefore, the pre-trained positional embeddings are interpolated (2D bicubic interpolation is adopted) to the sizes of the template and search region embeddings separately, which are further added to the patch embeddings.

\begin{figure}[t]
\centering
\setlength{\abovecaptionskip}{0.05cm}
\includegraphics[width=0.8\columnwidth, keepaspectratio]{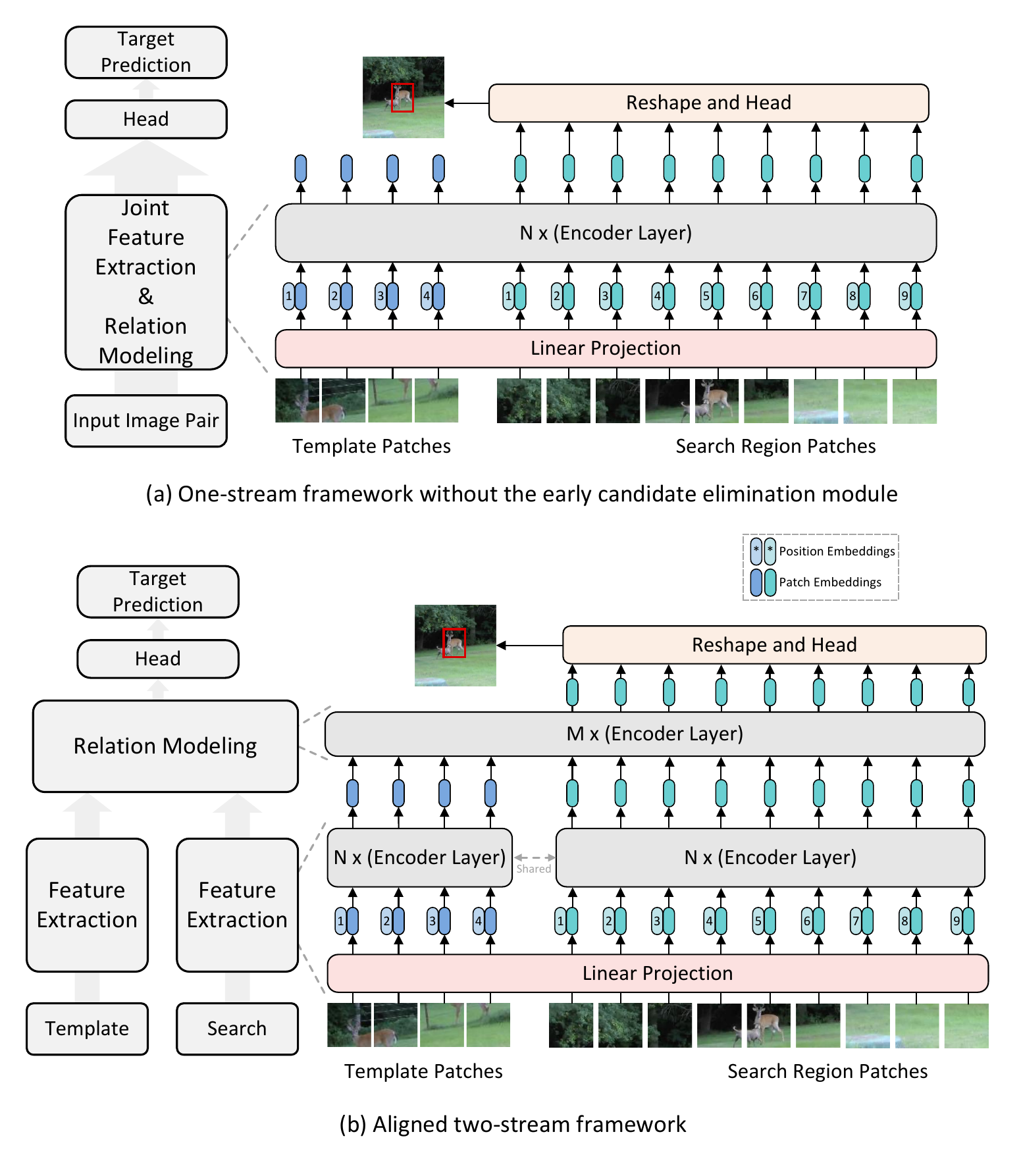}
\caption{\textbf{(a)} Our proposed one-stream framework without the early candidate elimination module, which combines feature extraction and relation modeling modules. \textbf{(b)} The aligned  two-stream tracking framework, which extracts features of the template and search region separately and then models the feature relation with extra Transformer encoder layers.}
\label{fig:arch_comapre}
\end{figure}

\textbf{Model Details.}
In Sec.~\ref{subsec:ablation}, we compare our OSTrack (without the early candidate elimination module) with aligned two-stream trackers (\ie, STARK-aligned and SwinTrack-aligned), and we further present the detailed structures in this section. The proposed one-stream framework, as shown in Fig.~\ref{fig:arch_comapre}(a), combines feature extraction and relation modeling modules into a single ViT backbone. The aligned two-stream framework,  as shown in Fig.~\ref{fig:arch_comapre}(b), first extracts features of the template and the search region separately with the same ViT backbone and then models the feature relation with several extra Transformer encoder layers. As presented in Sec.~\ref{subsec:ablation}, this relation modeling module is instantiated with the encoder structure proposed in STARK~\cite{stark} (STARK-aligned) and SwinTrack~\cite{swintrack} (SwinTrack-aligned) separately.

\textbf{Discriminative Regions Visualization.}
We show the method used to obtain the visualization of the discriminative regions in Fig.~\ref{fig:motivation}. Zagoruyko \etal~\cite{zagoruyko2017paying} show that the importance of a hidden neuron activation can be indicated by its absolute value. In this work, we adopt a similar approach to obtain the discriminative regions of each feature map $\boldsymbol{F}$. We first calculate the absolute mean values of each pixel along the channel dimension: 

\begin{equation}
M(\boldsymbol{F})=\frac{1}{C} \cdot \sum_{c=1}^C\left|\boldsymbol{F}_c\right|,
\end{equation}
where $C$ is the number of channels. Then, the relative importance of each pixel is then calculated by:

\begin{equation}
D(\boldsymbol{F})= \frac{M( \boldsymbol{F}) - \min(M(\boldsymbol{F})) }{\max(M(\boldsymbol{F})) - \min(M(\boldsymbol{F}))} ,
\end{equation}
where $ \min(\cdot) $ and $ \max(\cdot) $ return the maximum and minimum values of all pixels, respectively.

\section{More Ablation Studies}

\subsection{The Effect of Different Template Token Choices.}
As pointed out in Sec.~\ref{subsec:token_drop}, the goal of the early candidate elimination module is to identify and discard candidates belonging to background regions based on the ranking of similarity between the target and each candidate. However, the input template also contains background regions, which introduces noisy information when calculating the similarity score. Therefore,  different choices of template parts (tokens) used for the similarity calculation may influence the candidate elimination results and consequently affect the tracking performance. We compare four different template token choices (the similarity scores of all chosen template tokens are summed up for the final ranking): 1) all template tokens; 2) all template tokens within the ground truth target bounding box; 3) template tokens within a 4x4 region around the center of the template image; 4) the template token corresponding to the center of the template image. The result comparison of these template choices is shown in Tab.~\ref{tab:token_sec}. The results demonstrate that different template token choices do affect the quality of identifying background candidates. Since the input template contains background regions, directly using ``All Template Tokens'' clearly degrades the tracking performance compared with the baseline (``No Early Candidate Elimination''), \ie, 0.6\% lower in LaSOT AUC. Compared to other choices, using the central template token shows better performance, probably because the central token does not contain any background region and has aggregated the entire target information through self-attention.

\begin{table}[t]
\centering
\caption{Ablation study on different choices of template tokens used to identify candidates belonging to background.}
\label{tab:token_sec}
\resizebox{0.8\linewidth}{!}{
\begin{tabular}{c|ccc|ccc|ccc}
\hline
\multirow{2}{*}{Template   Token Selection} & \multicolumn{3}{c|}{LaSOT} & \multicolumn{3}{c|}{TrackingNet} & \multicolumn{3}{c}{GOT-10k} \\ \cline{2-10} 
                                       & Success & P$_{Norm}$ & P    & Success & P$_{Norm}$ & P    & AO    & SR$_{0.5}$ & SR$_{0.75}$ \\ \hline
No Early Candidate Elimination         & 68.7    & 78.1   & 74.6 & 82.9    & 87.5   & 81.6 & \best{73.6} & 83.0   & 71.7   \\
All Template Tokens                & 68.1    & 77.4   & 73.5 & 82.8    & 87.5   & 81.6 & 72.9 & 82.3   & 70.1   \\
All Template Tokens within GT Box & 68.5    & 78.1   & 74.2 & \best{83.1}    & \best{87.8}   & 81.7 & \best{73.6} & \best{83.4}   & \best{72.0}   \\
Center 4x4 Template Tokens         & 68.3    & 77.6   & 73.9 & 82.9    & 87.7   & \best{82.0} & 73.5 & 83.0   & 71.5   \\
Center Template Token              & \best{69.1}    & \best{78.7}   & \best{75.2} & \best{83.1}    & \best{87.8}   & \best{82.0} & \best{73.6} & 82.8   & 71.4   \\ \hline
\end{tabular}}
\end{table}

\subsection{Identity Embeddings and Relative Positional Embeddings}
We additionally verify the effect of adding identity embeddings and relative positional embeddings. Specifically, for the identity embeddings, we add learnable identity embeddings (to indicate a token belonging to the template or search region as in BERT~\cite{bert}) to template tokens and search region tokens separately. For the relative positional embeddings, the same method as in SwinTrack~\cite{swintrack} is adopted. The results are presented in Tab.~\ref{tab:rpe}, these two components do not bring performance gain compared to the original design, thus not adopted in our model.

\begin{table}[t]
\caption{The effect of adding additional identity embeddings to the template and search region embeddings and adding relative positional embeddings to the OSTrack-256 (without the early candidate elimination module). The results on LaSOT~\cite{lasot}, TrackingNet~\cite{trackingnet} and GOT10k~\cite{got10k} benchmarks are presented.}
\centering
\label{tab:rpe}
\resizebox{0.9\linewidth}{!}{
\begin{tabular}{l|ccc|ccc|ccc}
\hline
\multirow{2}{*}{}              & \multicolumn{3}{c|}{LaSOT} & \multicolumn{3}{c|}{TrackingNet} & \multicolumn{3}{c}{GOT-10k} \\ \cline{2-10} 
                    & Success & P$_{Norm}$ & P    & Success & P$_{Norm}$ & P    & AO    & SR$_{0.5}$ & SR$_{0.75}$ \\ \hline
Ours                & \best{68.7}    & \best{78.1}   & \best{74.6} & 82.9    & 87.5   & 81.6 & 73.6 & 83.0   & \best{71.7}   \\
+ Identity Embeddings & 68.0    & 77.3   & 73.6 & \best{83.3}    & \best{88.0}   & \best{82.2} & 73.6 & 82.9   & \best{71.7}   \\
+ Relative Positional Embeddings & 68.5    & 77.8    & 74.1   & 83.2      & 87.8      & 82.0     & \best{73.7}    & \best{83.3}    & 71.2   \\ \hline
\end{tabular}}
\end{table}

\subsection{Additional Relation Modeling Module}
To investigate whether our one-stream framework does not require an extra feature relation module, we add an additional transformer-based feature fusion module proposed in~\cite{swintrack}, which consists of 4 self-attention layers and 1 cross-attention layer, to further fusion the extracted template and search region features. As the results in Tab.~\ref{tab:add_relation} show, adding such a relation modeling module instead degrades the tracking performance, indicating that the output search region features of the ViT backbone have been sufficiently fused with the template features.

\begin{table}[t]
\caption{Add additional relation modeling module to our OSTrack-256 (without the early candidate elimination module).}
\centering
\label{tab:add_relation}
\resizebox{0.9\linewidth}{!}{
\begin{tabular}{l|ccc|ccc|ccc}
\hline
\multicolumn{1}{c|}{\multirow{2}{*}{}} & \multicolumn{3}{c|}{LaSOT} & \multicolumn{3}{c|}{TrackingNet} & \multicolumn{3}{c}{GOT-10k} \\ \cline{2-10} 
\multicolumn{1}{c|}{} & Success & P$_{Norm}$ & P    & Success & P$_{Norm}$ & P    & AO    & SR$_{0.5}$ & SR$_{0.75}$ \\ \hline
Ours                  & \best{68.7}    & \best{78.1}   & \best{74.6} & \best{82.9}    & \best{87.5}   & \best{81.6} & \best{73.6} & \best{83.0}   & \best{71.7}   \\
+ Relation Modeling     &  68.5    & 78.0   & 74.1 & \best{82.9}    & 87.4   & 81.5 & 72.7 & 82.2   & 70.5   \\ \hline
\end{tabular}}
\end{table}

\subsection{Fewer Relation Modeling Layers}
In the implementation of vanilla OSTrack, all encoder layers in ViT-Base (12 layers in total) are used for simultaneous feature extraction and relation modeling. In this subsection, we try to decrease the number of layers used for relation modeling. Specifically, only the last $n$ encoder layers are used for  simultaneous feature extraction and relation modeling, and the first $12 - n$ layers are only used for the template and search region feature extraction. $n$ is set to be 6 and 3 separately and the results are presented in Tab.~\ref{tab:fusion_ratio}. The results show that using fewer encoder layers for simultaneous feature extraction and relation modeling will degrade the tracking performance, showing the necessity of sufficient feature fusion.

\begin{table}[t]
\caption{Ablation studies on the number of encoder layers used for relation modeling.}
\centering
\resizebox{0.7\linewidth}{!}{
\begin{tabular}{c|ccc|ccc|ccc}
\hline
\multirow{2}{*}{} & \multicolumn{3}{c|}{LaSOT} & \multicolumn{3}{c|}{TrackingNet} & \multicolumn{3}{c}{GOT-10k} \\ \cline{2-10} 
    & Success & P$_{Norm}$ & P    & Success & P$_{Norm}$ & P    & AO    & SR$_{0.5}$ & SR$_{0.75}$ \\ \hline
12 (Ours) & \best{68.7}    & \best{78.1}   & \best{74.6} & 82.9    & \best{87.5}   & \best{81.6} & \best{73.6} & \best{83.0}   & \best{71.7}   \\
6  & 67.9    & 77.3   & 73.6 & \best{83.0}    & \best{87.5}   & 81.5 & 73.3 & 82.9   & 71.4   \\
3  & 67.8    & 77.0   & 73.5 & 82.7    & 87.4   & 80.7 & 72.8 & 82.5   & 70.7   \\ \hline
\end{tabular}}
\label{tab:fusion_ratio}
\end{table}

\subsection{Different Token Drop Rate}
We also try to apply a different keeping ratio $\rho$ for the early candidate elimination module. As the results in Tab.~\ref{tab:keep_ratio} show, using $\rho < 0.7$ leads to performance drop on the LaSOT~\cite{lasot} tracking benchmark since small $\rho$ may cause a significant information loss. However, the reduction in computational cost that comes with large $\rho$ is limited. Setting $\rho = 0.7$ shows a decent decrease in computational cost with a slight improvement in tracking performance. Therefore, we use $\rho = 0.7$ in our experiments.

\begin{table}[t]
\caption{Different keeping ratio $\rho$ used in the early candidate elimination module ($\rho = 1$ means the early candidate elimination module is not adopted).}
\centering
\label{tab:keep_ratio}
\resizebox{0.9\linewidth}{!}{
\begin{tabular}{c|ccc|ccc|ccc|c}
\hline
\multirow{2}{*}{\begin{tabular}[c]{@{}c@{}}Keeping \\ Ratio\end{tabular}} & \multicolumn{3}{c|}{LaSOT} & \multicolumn{3}{c|}{TrackingNet} & \multicolumn{3}{c|}{GOT-10k} & \multirow{2}{*}{MACs (G)} \\ \cline{2-10}
    & Success & P$_{Norm}$ & P    & Success & P$_{Norm}$ & P    & AO    & SR$_{0.5}$ & SR$_{0.75}$ &      \\ \hline
1   & 68.7    & 78.1   & 74.6 & 82.9    & 87.5   & 81.6 & 73.6 & 83.0   & 71.7   & 29.0   \\ \hline
0.9 & 68.7    & 78.2   & 74.6 & 83.2    & 87.8   & 82.0 & \best{74.1} & \best{83.6}   & \best{71.8}   & 26.2 \\
0.8 & 68.9    & 78.4   & 74.9 & \best{83.3}    & \best{88.0}   & \best{82.3} & 73.4 & 82.7   & 71.4   & 23.6 \\
0.7 & \best{69.1}    & \best{78.7}   & \best{75.2} & 83.1    & 87.8   & 82.0 & 73.6 & 82.8   & 71.4   & 21.5 \\
0.6 & 68.4    & 77.9   & 74.3 & 83.1    & 87.6   & 81.8 & 73.5 & 82.9   & 71.7   & 19.6 \\
0.5 & 67.8    & 77.2   & 73.4 & 82.7    & 87.4   & 81.3 & 71.8 & 81.3   & 68.4   & \best{18.0}   \\ \hline
\end{tabular}}
\end{table}

\section{Results on VOT2020}
VOT2020~\cite{vot2020} is a challenging short-term tracking benchmark that is evaluated by target segmentation results. To evaluate OSTrack on VOT2020, we use AlphaRefine~\cite{alpha_refine} to generate segmentation masks, and the results are shown in Tab.~\ref{tab:vot20_results}. Since the wide existence of distractors in VOT2020, updating the template during the tracking process has become a common practice to avoid tracking drift, which can bring significant performance gain (\eg, STARK-ST50~cite{stark} raises the EAO from 0.462 to 0.505 by simply adding a dynamic template).  
OSTrack-256 obtains an EAO of 0.518, which already outperforms the STARK-ST50 with an online template updating mechanism. This demonstrates the great potential of OSTrack which serves as a neat and strong baseline model. 

\begin{table}[t!]
\caption{Comparison on VOT2020 benchmark. The left part of the trackers adopts an online template update mechanism, while the right part of the trackers does not. The best two results are shown in \rankfirst{red} and \ranksecond{blue} fonts.}
\centering
\label{tab:vot20_results}
\resizebox{\linewidth}{!}{
\begin{tabular}{c|cccccc|cccc}
\hline
 &
  \begin{tabular}[c]{@{}c@{}}Ocean\\ ~\cite{ocean}\end{tabular} &
  \begin{tabular}[c]{@{}c@{}}ATOM\\ ~\cite{atom}\end{tabular} &
  \begin{tabular}[c]{@{}c@{}}D3S\\ ~\cite{d3s}\end{tabular} &
  \begin{tabular}[c]{@{}c@{}}AlphaRef\\ ~\cite{alpha_refine}\end{tabular} &
  \begin{tabular}[c]{@{}c@{}}STARK-\\ ST50~\cite{stark}\end{tabular} &
  \begin{tabular}[c]{@{}c@{}}STARK -\\ ST101~\cite{stark}\end{tabular} &
  \begin{tabular}[c]{@{}c@{}}SiamMask\\ ~\cite{siammask}\end{tabular} &
  \begin{tabular}[c]{@{}c@{}}STARK-\\ S50~\cite{stark}\end{tabular} &
  OSTrack-256 &
  OSTrack-384 \\ \hline
EAO ($\uparrow$)        & 0.43  & 0.271 & 0.439 & 0.482 & 0.505 & 0.497 & 0.321 & 0.462 & \ranksecond{0.518} & \rankfirst{0.524} \\
Accuracy ($\uparrow$)   & 0.693 & 0.462 & 0.699 & 0.754 & 0.759 & \ranksecond{0.763} & 0.624 & 0.761 & 0.762 & \rankfirst{0.767} \\
Robustness ($\uparrow$) & 0.754 & 0.734 & 0.769 & 0.777 & 0.817 & 0.789 & 0.648 & 0.749 & \ranksecond{0.814} & \rankfirst{0.816} \\ \hline
\end{tabular}
}
\end{table}

\section{Results on ITB}
ITB~\cite{itb} benchmark is a newly collected benchmark with 9 representative scenarios and 180 diverse videos, which contains more informative tracking sequences. Tab.~\ref{tab:itb_results} shows the results of OSTrack compared with other SOTA tackers. Our OSTrack-384 achieves 64.8\% in mIoU, surpassing the previous best tracker STARK~\cite{stark} by a large margin (7.2\%).

\begin{table}[t!]
\centering
\caption{Comparison with state-of-the-arts on ITB~\cite{itb} benchmark. mIoU(\%) scores are reported. The best two results are shown in  \rankfirst{red} and \ranksecond{blue} fonts.}
\resizebox{\linewidth}{!}{
\begin{tabular}{c|cccccccccc|cc}
\hline
 &  \begin{tabular}[c]{@{}c@{}}SiamRPN++\\ \cite{siamrpn++}\end{tabular}
 & \begin{tabular}[c]{@{}c@{}} Ocean\\ \cite{ocean}\end{tabular}
 & \begin{tabular}[c]{@{}c@{}}GAT\\ \cite{gat}\end{tabular}
 & \begin{tabular}[c]{@{}c@{}}ATOM\\ \cite{atom}\end{tabular}
 & \begin{tabular}[c]{@{}c@{}}DiMP\\ \cite{dimp}\end{tabular} 
 & \begin{tabular}[c]{@{}c@{}}PrDiMP\\ \cite{prdimp}\end{tabular} 
 & \begin{tabular}[c]{@{}c@{}}KYS\\ \cite{kys}\end{tabular} 
 & \begin{tabular}[c]{@{}c@{}}TrDiMP\\ \cite{trdimp}\end{tabular} 
 & \begin{tabular}[c]{@{}c@{}}TransT\\ \cite{transt}\end{tabular}
  & \begin{tabular}[c]{@{}c@{}}STARK\\ \cite{stark}\end{tabular}  
 & \begin{tabular}[c]{@{}c@{}}OSTrack\\ -246\end{tabular} 
 & \begin{tabular}[c]{@{}c@{}}OSTrack\\ -384\end{tabular} \\ \hline
mIoU    & 44.1 & 47.7 & 44.9 & 47.2  & 53.7 & 54.4 & 52.0    & 56.1 & 54.7  & 57.6 & \ranksecond{61.2} & \rankfirst{64.8} \\ \hline

\end{tabular}}

\label{tab:itb_results}
\end{table}

\section{More Visualization}
\begin{figure}[b!]
\centering
\setlength{\abovecaptionskip}{0.05cm}
\includegraphics[width=0.7\columnwidth, keepaspectratio]{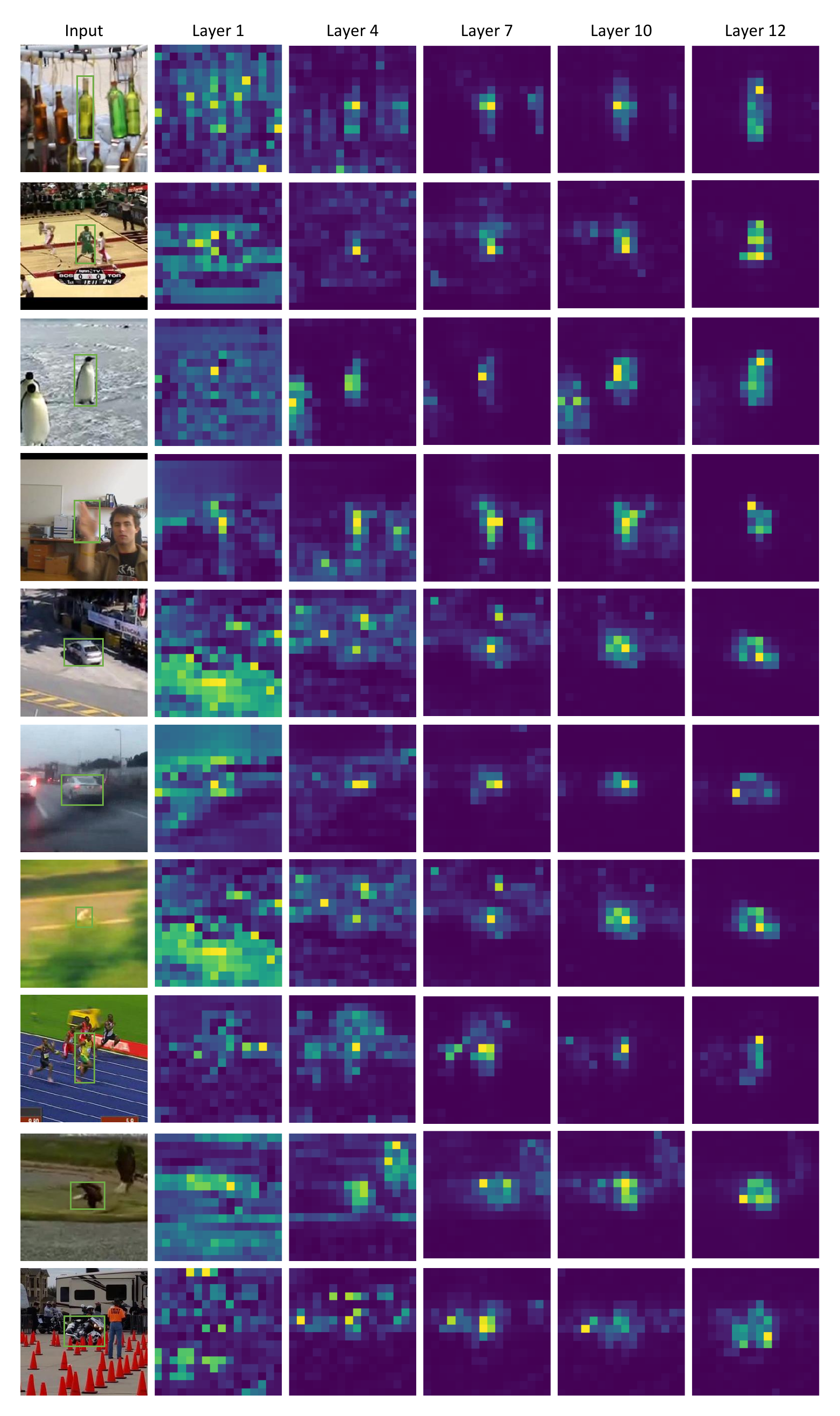}
\caption{Extended visualization for attention weights of the search region corresponding to the center part of the template after different ViT layers, the \textcolor{green}{green} rectangles indicate target objects. The results show that our one-stream framework is able to distinguish between targets and distractors and progressively focus on targets.}
\label{fig:vis_attn_more}
\end{figure}

We first provide more visualization results for attention weights of the search region corresponding to the center part of the template (which can be seen as the target) in Fig.~\ref{fig:vis_attn_more}. The results show that the model attends to the foreground objects at an early stage (see ``Layer 4'' in Fig.~\ref{fig:vis_attn_more}) and finally shows great discriminative power between the target and distractors (see ``Layer 12'' in Fig.~\ref{fig:vis_attn_more}). These phenomenons demonstrate that the proposed OSTrack can extract target-oriented features with strong target-distractor discriminability.

\begin{figure}[t]
\centering
\setlength{\abovecaptionskip}{0.05cm}
\includegraphics[width=\columnwidth, keepaspectratio]{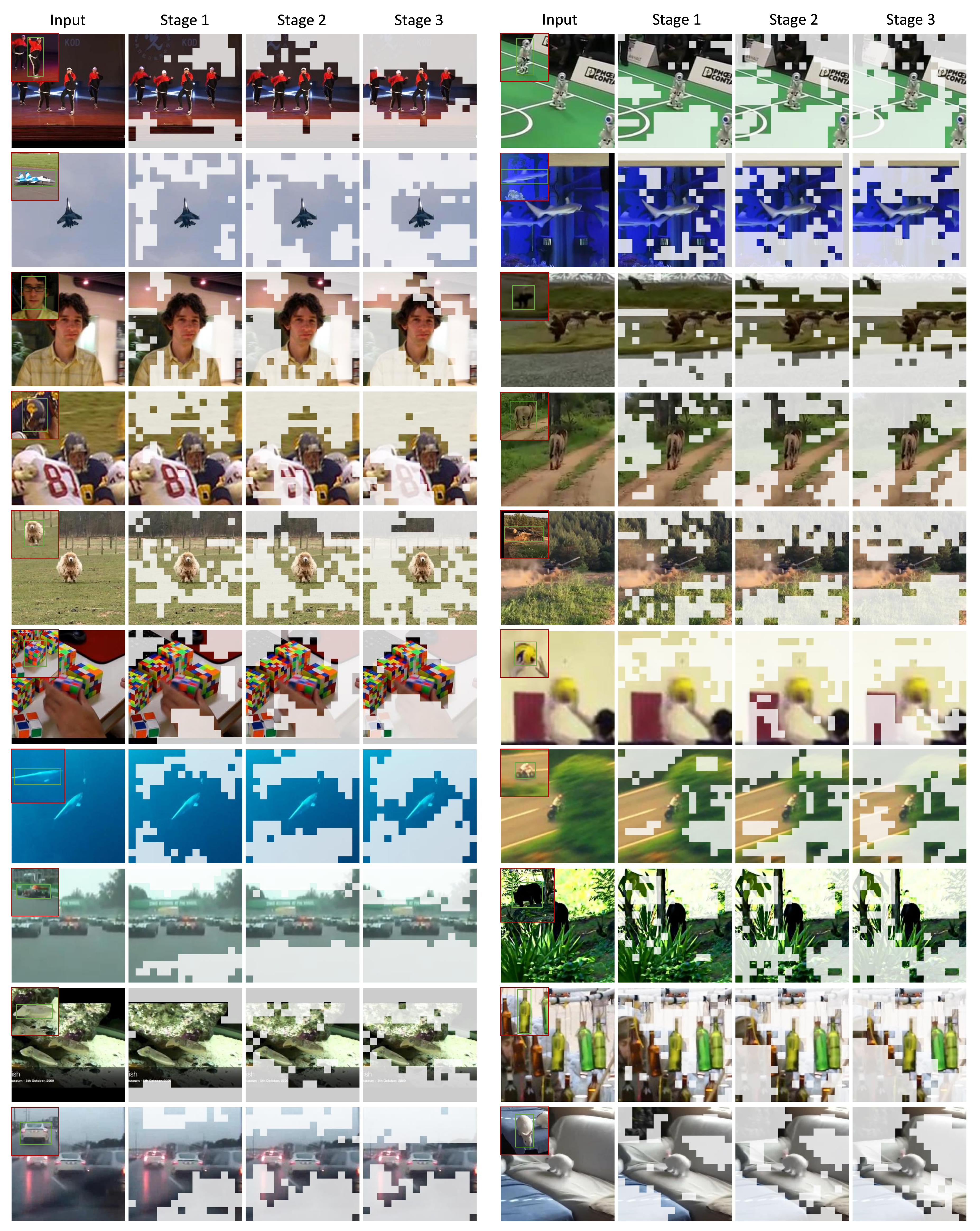}
\caption{Extended visualization results of the progressive candidate elimination process. The main body of ``Input'' is the search region image, and the upper left corner shows the corresponding template image. The \textcolor{green}{Green} rectangles indicate target objects and the masked regions represent the discarded tokens. Our early candidate elimination module can effectively deal with different tracking targets and scenarios.}
\label{fig:vis_drop_more}
\end{figure}

In Fig.~\ref{fig:vis_drop_more}, more visualization results of the early candidate elimination module are presented. The results validate that the proposed method can effectively identify and discard background regions under various target categories and challenge scenarios (\eg, target deformation, occlusion, motion blur, \etc).

\end{document}